\documentclass[conference]{IEEEtran}
\usepackage{times}
\usepackage{booktabs}
\usepackage{amsmath}
\usepackage{amsfonts}
\usepackage{amssymb}
\usepackage{graphicx}
\usepackage[numbers]{natbib}
\usepackage{multicol}
\usepackage[bookmarks=true]{hyperref}
\usepackage{lipsum}

\begin{document}

\title{RobotFleet: An Open-Source Framework for Centralized Multi-Robot Task Planning}

\author{\authorblockN{Rohan Gupta$^*$, Trevor Asbery$^*$, Zain Merchant$^*$, Abrar Anwar, Jesse Thomason}
\authorblockA{University of Southern California}
}



%

\maketitle

\begin{abstract}
Coordinating heterogeneous robot fleets to achieve multiple goals is challenging in multi-robot systems.
We introduce an open-source and extensible framework for centralized multi-robot task planning and scheduling that leverages LLMs to enable fleets of heterogeneous robots to accomplish multiple tasks.
RobotFleet provides abstractions for planning, scheduling, and execution across robots deployed as containerized services to simplify fleet scaling and management.
The framework maintains a shared declarative world state and two-way communication for task execution and replanning. 
By modularizing each layer of the autonomy stack and using LLMs for open-world reasoning, RobotFleet lowers the barrier to building scalable multi-robot systems.
The code can be found here: \href{https://github.com/therohangupta/robot-fleet}{github.com/therohangupta/robot-fleet}.
\end{abstract}

\IEEEpeerreviewmaketitle

\section{Introduction}
Heterogeneous robots, ranging from mobile manipulators to navigation-focused platforms, are challenging to coordinate for solving a diverse set of goals.
Differences in capabilities and communication interfaces complicate fleet-level planning and allocation, particularly as both the number of robots and the number of goals grow.
Effectively coordinating such fleets requires more than just optimal task planning and allocation; it also needs scalable infrastructure for planning, scheduling, execution, and replanning in dynamic environments.

Past work has considered planning for single robots~\cite{ahn2022can,huang2022inner,singh2023progprompt,anwar2024remembr,shah2023lm}, coordinating multiple manipulators~\cite{mandi2024roco,zhang2024towards,chu2025llm}, and multi-robot task planning~\cite{wang2024dart,bai2024twostep,kannan2024smart,liu2024coherent,li2025large,obata2024lip,sikand2021robofleet,chen2024scalable,wang2024mosaic}. 
However, these approaches assume fixed capabilities or produce only simulated results.
They do not address the systems-level difficulties involved in practically deploying and managing heterogeneous fleets with varying capabilities given multiple goals.
To effectively deploy multi-robot systems, we must scale to new goals, capabilities, and robots.

We present RobotFleet, an open-source, extensible framework for centralized multi-robot task planning and scheduling.
RobotFleet leverages large language models (LLMs) to construct dependency graphs from high-level goals, allocate subtasks across a heterogeneous fleet using a centralized scheduler, and coordinate task execution across containerized robot workers. 
By containerizing the robot workers, we are able to easily scale to new robots.
The framework maintains a declarative world state, supports two-way communication for task updates, and modularizes the autonomy stack to enable flexible replanning.

RobotFleet is designed to simplify the deployment of real-world multi-robot fleets.
Users can register new robots through a command line interface (CLI), define custom task planning or allocation logic, and modify the world representation without needing to manage low-level communication or coordination overhead.
Our hope is that this abstraction will allow researchers and developers to focus on developing system behaviors while RobotFleet handles the complexity of distributed robot execution.

\begin{figure}
    \centering
    \includegraphics[width=\linewidth]{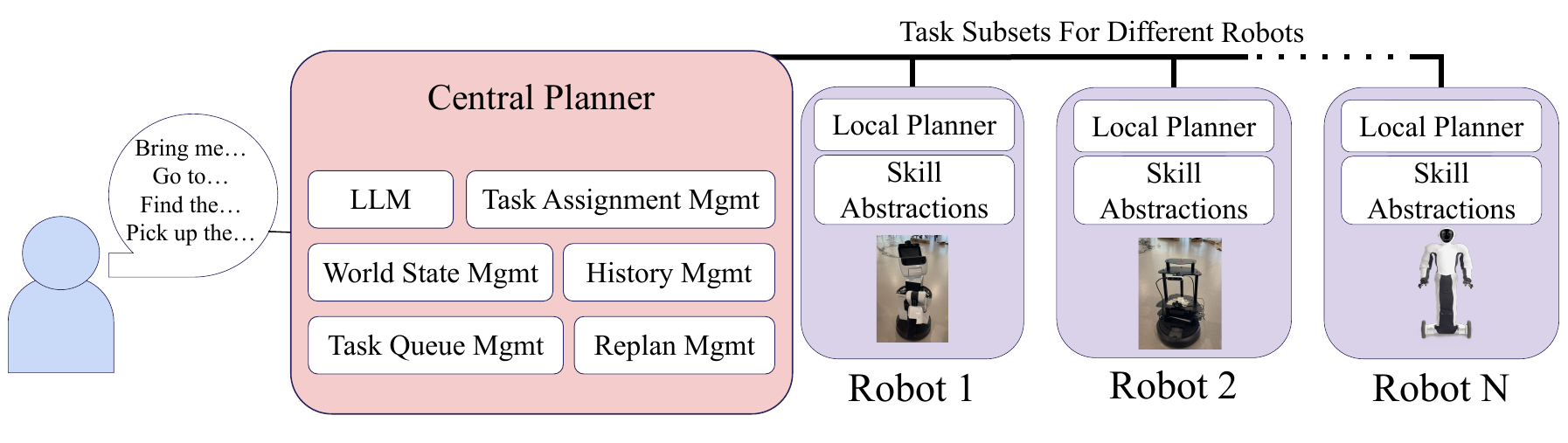}
    \caption{\textbf{RobotFleet} is an open-source framework for coordinating multiple goals across a fleet of heterogeneous robots by leveraging LLMs for task planning and allocation.}
    \label{fig:placeholder}
\end{figure}


\begin{figure*}[t]
    \centering
    \includegraphics[width=.9\linewidth]{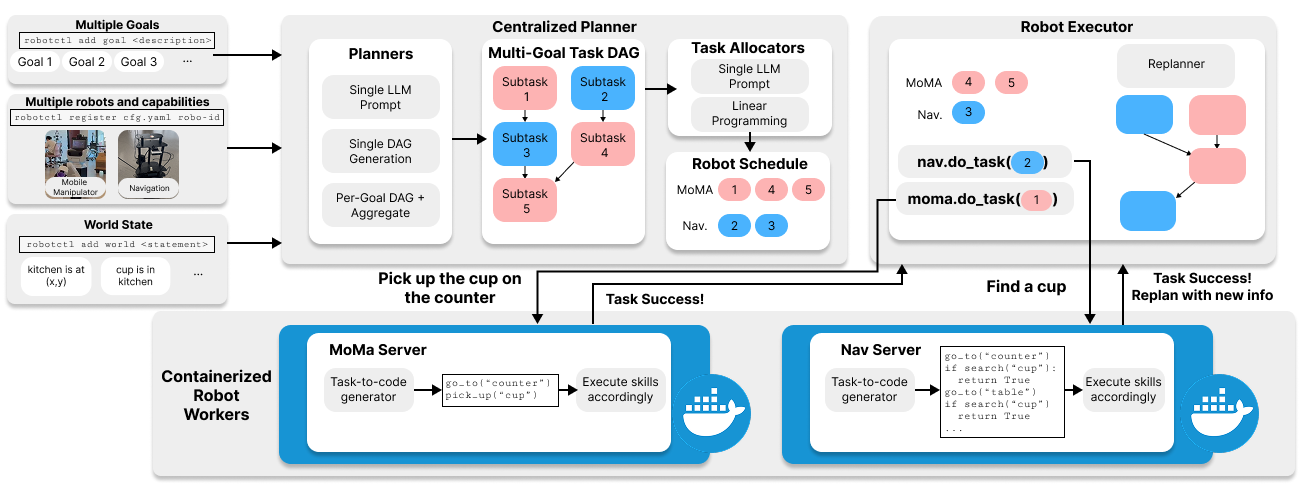}
    \caption{\textbf{Overview.} RobotFleet consists of three phases: task planning from goals, task allocation to robots, and task execution on optionally-containerized robot workers. For task planning, we support multiple types of task planners which generate plans in the form of DAGs. These tasks in the generated plans are then allocated to and executed across multiple robots.}
    \label{fig:fig1}
\end{figure*}

\section{Related Work}

Task planning for single robots has long focused on generating action sequences that allow an agent to achieve predefined goals under environmental and capability constraints. 
Classical approaches, such as symbolic planners like STRIPS~\cite{fikes1971strips}, PDDL~\cite{fox2003pddl2}, or ASP~\cite{lifschitz2019answer} operate under assumptions of full-state observability, but provide guarantees of success when a solution exists. 
Probabilistic models and decision-theoretic planners have been introduced to handle uncertainty and partial observability~\cite{kushmerick1995algorithm, boutilier1999decision}. 
Large language models (LLMs) have been leveraged for high-level reasoning and plan generation, trading flexible plans and semantic understanding in unstructured domains~\cite{DBLP:journals/corr/abs-2201-07207, 10161317, rana2023sayplangroundinglargelanguage} at the cost of plan fidelity.

Multi-robot task planning introduces additional challenges such as coordination, communication, and task allocation across heterogeneous agents. 
Past work has explored both centralized and decentralized paradigms~\cite{li2025large, chen2024scalable}. 
Centralized systems can exploit global knowledge to optimize plans, but scalability and robustness under communication delays or failures remain concerns. 
Decentralized systems~\cite{chen2024scalable} typically trade optimality for scalability by relying on local decision-making and communication among agents. 
LLMs have recently been explored as tools for enabling high-level reasoning across robot fleets, supporting capabilities like task decomposition and low-level motion planning~\cite{kannan2024smart, 9013090, shiarlis2018tacolearningtaskdecomposition, yu2025conavgptmultirobotcooperativevisual}. 
Classical optimization methods, including linear programming, constraint satisfaction, and heuristic algorithms, have also been applied for better planning and allocation~\cite{bai2024twostep, chakraa2023optimization, obata2024lip}.
Research on multi-robot fleet systems has also explored the design of robust software architectures that facilitate distributed coordination and state management~\cite{sikand2021robofleet, guzman2021fleet}.
However, these prior works focused only on navigation and relied on rigid systems like ROS.
In this work, we build a more generic software architecture for multi-robot planning and allocation that is built to extend to arbitrary embodiments and tasks.

\section{RobotFleet}
We introduce RobotFleet, designed to be an extensible approach to building multi-robot systems. 
RobotFleet is designed around three key modules: (1) LLM-driven task planning, (2) centralized task allocation, and (3) distributed execution through containerized robot agents.

\textbf{System Design.}
The multi-robot system is designed with a hierarchical architecture as shown in Figure~\ref{fig:fig1} that enables coordinated task execution across a heterogeneous fleet of robots. 
At its core, the system employs a centralized fleet manager that orchestrates robot operations through a planning and allocation framework. 
Robots are registered with specific capabilities and are deployed as containerized services for easy management and scalability. 
Though our framework enables any approach to accomplish tasks, once a robot is given a natural language task through the scheduler, we generate code to call robot skills, similar to past work~\cite{singh2023progprompt,liang2023code}.
The framework maintains a world state through declarative statements about the environment and robot positions which are used to inform planning decisions. 
The framework supports dynamic task management, allowing for real-time updates to the world state and task status, while maintaining a clear separation between planning, allocation, and execution phases. 

\textbf{Containerized Deployment.}
To support heterogeneous robots and promote system scalability, RobotFleet can work with Docker-based containerization for robot deployment. 
Containerization is not explicitly enforced as robots can work with the central planner as long as they specify a host and port for communication. 
For example, in our demo in Section~\ref{realdemo}, the Human Support Robot (HSR) was running on a Docker container with a ROS2 Humble configuration, while our LoCoBot ran its standard Ubuntu runtime environment in ROS1, and both were able to run in real-time. 
Both containerization and running on the default non-container runtime environment of the robot are supported.
We leave it up to the user to decide whether containerization makes sense for their robot setup.  

Running and updating robot runtime environments with containers will be useful when there are many robots of the same type in a fleet. 
Containerized deployment can be handled by a separate CI/CD pipeline, where if changes are made to the environment of a specific type of robot, the system can quickly and easily deploy that environment to all robots of a specific type and launch it. 

During registration, each robot is configured with its capabilities in a YAML file that is provided by the user as well as deployment details such as a container image.
These settings are stored by the fleet manager and can be utilized to manage different robots' environments simultaneously with Docker containers. 

\textbf{Modularity.}
RobotFleet is designed to be modular and open-source under the MIT license, allowing researchers and developers to experiment with different components of the system. 
For example, the planner, task allocator, or execution engine can each be swapped out independently without overhauling the entire system.
Next, we discuss our implementations for planning and allocation.

\subsection{Task Planning}

Before assigning tasks to individual robots, we represent the overall mission (set of goals) as a directed acyclic graph (DAG) to capture dependencies between subtasks similar to past work~\cite{obata2024lip,wang2024dart}. 
Let \(G = \{g_1, \dots, g_k\}\) be the set of high-level goals. 
Each goal is decomposed into a sequence of subtasks \(T = \{t_1, \dots, t_n\}\), with dependencies encoded in the DAG structure. 
We support three planning strategies to construct this DAG, each using an LLM to generate structured task plans.

\textbf{\texttt{Per‐Goal DAG}}: In this mode, we ask the LLM\footnote{We used GPT4 \cite{openai2024gpt4technicalreport} in our experiments} to generate a separate DAG \(\mathcal{D}_\ell = (V_\ell, E_\ell)\) for each goal \(g_\ell\). The choice of LLM is customizable per user preference — the prompts are intended to be transferable across different LLMs with minimal further prompt finetuning. 
This approach separates out each goal into separate DAGs, and may perform better if goals have minimal shared subtasks.

\textbf{\texttt{Big‐DAG}}: Here, we request the LLM to produce a single DAG \({\mathcal{D}_{\rm big} = (V, E)}\) for all goals combined. 
Each node \({v_i \in V}\) corresponds to a subtask description, and each edge \({(v_i, v_j) \in E}\) encodes a dependency (i.e., \(t_i\) must precede \(t_j\)). 
The LLM is prompted to output explicit \texttt{depends\_on} relationships, which we then topologically sort to create the task execution order.

\textbf{\texttt{Monolithic Prompt}}: In this strategy, we concatenate all goals and world state information into a single prompt to the LLM. 
The model outputs a flat list of tasks \(\{(d_i, \Gamma_i)\}\), where \(d_i\) is the task description and \(\Gamma_i\) represents inferred dependencies. 
These tasks and dependencies are parsed into a task list and converted into a DAG based on the model’s inferred ordering.

The final output is transformed into a unified \texttt{Plan} format that standardizes the task descriptions and dependency graph before moving on to the allocation phase.

\subsection{Task Allocation}
\label{sec:allocation}
We have two approaches for assigning tasks \({T = \{t_1, \dots, t_n\}}\) to robots \({R = \{r_1, \dots, r_m\}}\): an LLM–based method and a simple mixed-integer linear program (MILP) that minimizes the maximum task load across all robots given capability constraints on tasks. If the set of tasks requires capabilities that none of the robots in the current fleet of registered robots has, then it will yield no allocation solution. 

\textbf{\texttt{LLM-Based Allocation.}}  
One approach for task allocation is to use LLMs.
Given a list of tasks \((t_i, d_i, \Gamma_i)\), robots \((r_j, \mathcal{C}_j)\), and world statements \(W\), we prompt the LLM to return a JSON mapping \(\{t_i: r_j\}\). 
The prompt includes all relevant task descriptions, robot capabilities, and environmental context. 
The model’s response defines an allocation map \(\mathcal{A}_{\text{LLM}}: T \rightarrow R\).

\textbf{\texttt{MILP-Based Allocation.}}  
We can also formulate task allocation as a mixed-integer linear program that minimizes the maximum task load \(M\) across all robots given capability constraints. 
Let \(x_{i,j} \in \{0,1\}\) indicate whether task \(t_i\) is assigned to robot \(r_j\). The core objective and load-balancing constraint are:

\begin{equation}
\min_{x, M} \; M \quad \text{s.t.} \quad \sum_i x_{i,j} \le M \;\; \forall j
\end{equation}

Each task is assigned to one robot: \({\sum_j x_{i,j} = 1 \; \forall i}\).
The capability constraints for each robot are derived from the capabilities, which are a set of strings in the robot's YAML file. The constraints for each task are extracted by manually parsing for keywords and processing them into a set of strings representing the needed capabilities. For example, if the task is ``Navigate to the kitchen and explore the area", then we can extract ``navigation" and ``exploration" as inferred capabilities for the task. If a task's capability constraints are a subset of a robot's capabilities, then the robot can be assigned to assign to the task by the LP. This is enforced by setting \({x_{i,j} = 0}\) when \({C_{t_i} \not\subseteq \mathcal{C}_{r_j}}\).

If the MILP is feasible, we apply the resulting assignment \({\mathcal{A}_{\text{LP}}(t_i) = r_j}\) using the optimal solution \({x^*_{i,j}}\). 
If no solution exists, we fall back to round-robin assignment.

\subsection{Replanning}

Once the robot is able to plan and allocate tasks to robots, it will often experience failures or need to propagate information back to the central planner. 
In our setup, a given robot will retry up to three times to complete a task.
If it does not succeed in the task, it will replan.
We maintain a connection to the central planner such that each robot can send replan requests.
On a replan, the central planner is able to update its world state prompt, and additionally use the current progress of the existing plan to generate a new plan and allocate tasks accordingly.

\section{User Experience and System Extensibility}

RobotFleet was designed in such a way that each component is modularized and can be extended to new versions of planners, allocators, executors, and robot task servers without making large refactors to the code. 
We define abstract base classes for each component and allow developers to create their own versions of each to see what works best in their fleet. 
We also encourage the use of saving important prompt information in a fleet rules directory for ease of prompt testing and version control. 
Finally, we use Docker containers and well-defined configurations to manage the fleet of robot containers; we believe this setup will be the best way to manage robots together at scale. 
The careful design of this system will allow roboticists to easily extend and use RobotFleet for their specific heterogeneous robot setup. 

\textbf{CLI Usage.}
As shown in left corner of Figure \ref{fig:fig1}, users can configure the information available to the robot fleet using simple CLI commands. 
We developed \texttt{robotctl} to be an easy CLI for users can register different robots using a robot config YAML file, which have a standard set of details such as robot name, capabilities, network connection information, and other details. 
Users can use \texttt{robotctl} to add world state statements in the form of natural language, goals for the robot fleet, and spawn different robots into the system based on config files.
Users can also manually add tasks to specific plans. 
The CLI interface makes it easy to add, remove, and configure the robot fleet to achieve multiple goals specified in natural language. 
This interface allows the system to be simply extended to accommodate another goal, robot, plan, task, or world state. 

\textbf{Robot Servers.}
Each robot, whether it is running a Docker container or just is on its standard runtime environment, starts a server that contains deployment information including the exposed task API and the Docker deployment configuration, if applicable. 
A Docker container would be composed of the specific dependencies that any user can define depending on their robot, such as using ROS, Python packages, or other systems to control their robot.

In our demo, our robots use ROS as the underlying system along with APIs built by the robots' original developers to control the robot, and a ProgPrompt-style~\cite{singh2023progprompt} code generation interface to handle language-defined instructions. 

\section{Testing}

\begin{table}[t]
\centering
\begin{tabular}{llrr}
\toprule
Planner         & Allocator       & \multicolumn{1}{c}{3 Robots/5 Goals} & \multicolumn{1}{c}{5 Robots/10 Goals} \\
\midrule
Monolithic Prompt & LP               & 81.1 ± \phantom{0}3.9                          & 60.0 ± \phantom{0}3.9                             \\
Monolithic Prompt & LLM              & 74.7 ± 21.8                         & 90.0 ± \phantom{0}3.9                             \\
Big DAG           & LP               & 37.6 ± 22.1                         & 60.0 ± 11.6                            \\
Big DAG           & LLM              & 46.2 ± \phantom{0}7.3                          & 52.5 ± 11.7                            \\
Per-Goal DAG      & LP               & 35.0 ± 18.1                         & 42.9 ± \phantom{0}8.8                             \\
Per-Goal DAG      & LLM              & 56.9 ± 12.5                         & 68.1 ± 17.6                            \\
\bottomrule
\end{tabular}
\caption{Idle time percentage for each planner–allocator pairing.  Introducing the DAG structure leads to lower idle times.}
\label{tab:idle-times}
\end{table}

\begin{figure*}[t]
    \centering
    \includegraphics[width=1.0\linewidth]{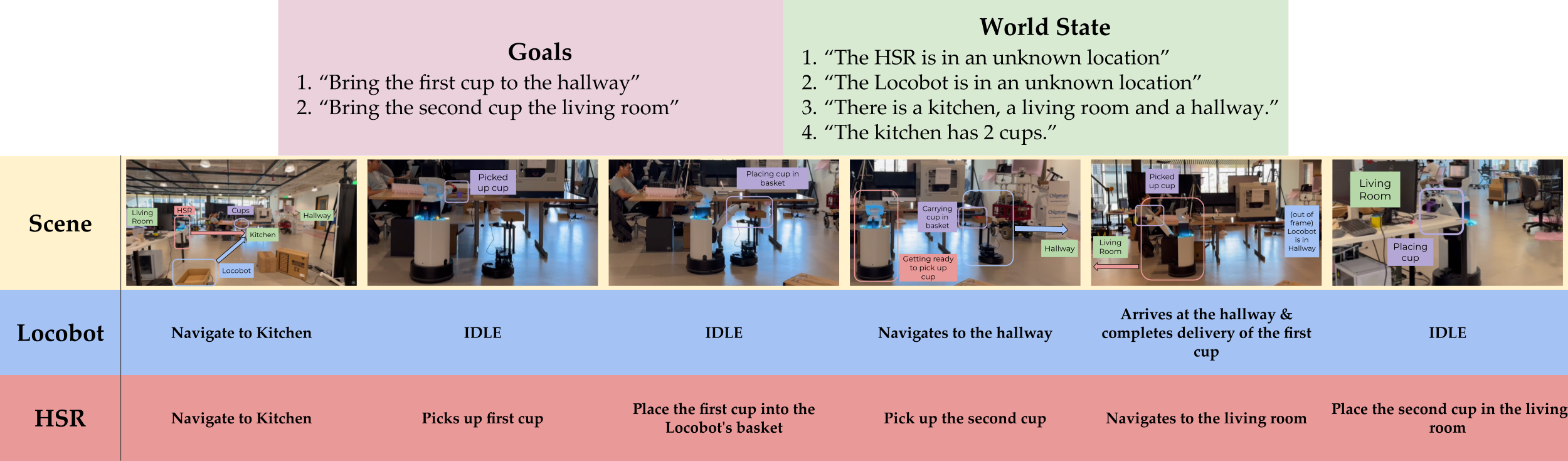}
    \caption{We show the end-to-end real-robot execution from RobotFleet based on two provided high-level goals, which were decomposed into dependent subtasks, as shown in the DAG in Figure~\ref{fig:fig4}. We use a Locobot navigation robot and an HSR mobile manipulator and are able to effectively separate subtasks to accomplish multiple goals.}
    \label{fig:fig3}
\end{figure*}

\begin{figure}[t]
    \centering
    \includegraphics[width=1.0\linewidth]{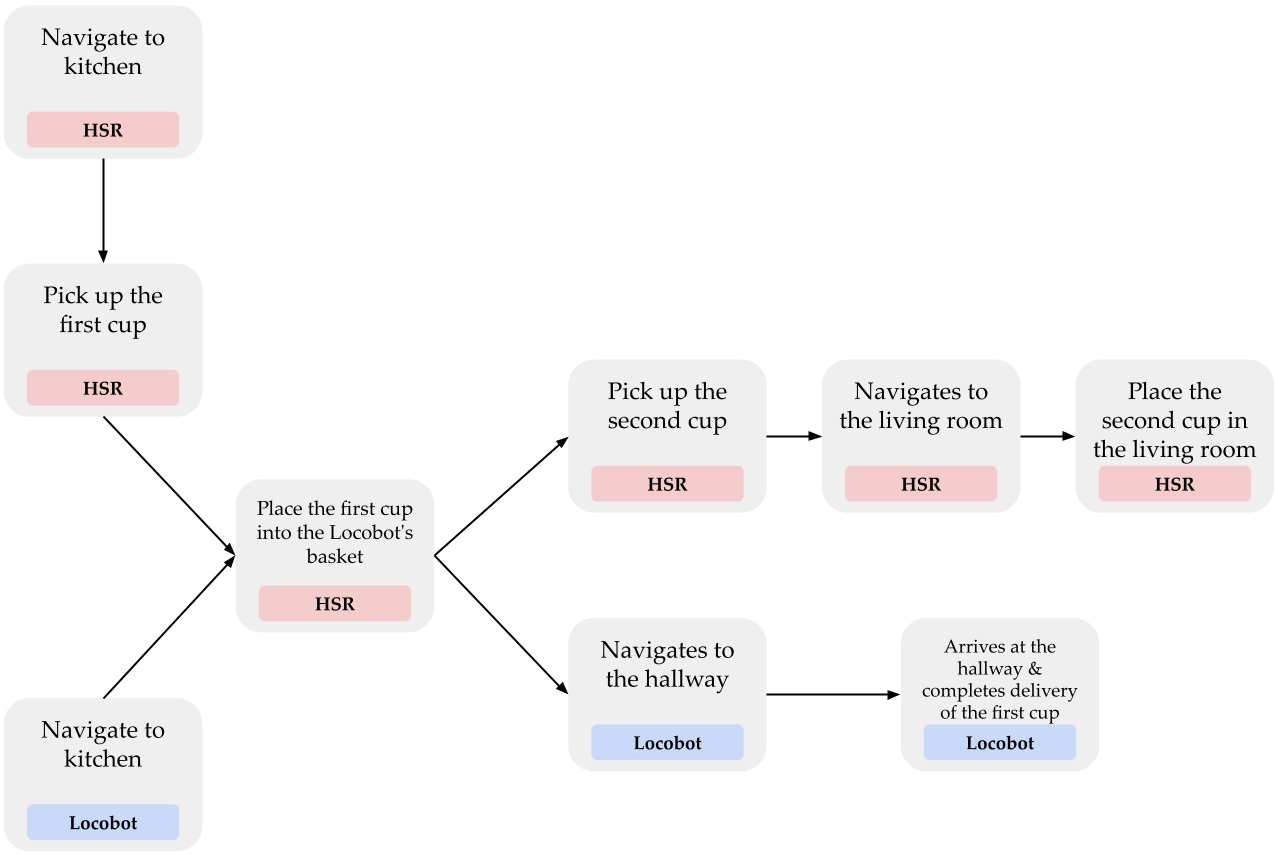}
    \caption{The directed acyclic graph (DAG) representing the task plan and their dependencies created by RobotFleet given the goals described in Figure~\ref{fig:fig3}. We find that the generated DAG can effectively decompose multiple general goals into dependent multi-robot subtasks.}
    \label{fig:fig4}
\end{figure}

We test our RobotFleet's task planning and scaling capabilities by simulating dozens of robots. We investigate the idle times of the robots during the execution of the generated plan. 
We also demonstrate the planning system by generating a plan and deploying it across two real robots. 

\subsection{Efficient Scaling.}
We simulate multiple goals across robots of varying capabilities to analyze how well our framework scales and evaluate how different planning and allocation strategies impact fleet utilization.
For each scenario, we instantiated RobotFleet with the specified number of robot instances and automatically generated human‐readable task descriptions corresponding to everyday activities such as ``make breakfast". 
Each experiment then used every combination of three planners (Monolithic Prompt, Big DAG, and Per-Goal DAG) and two allocators (Linear Programming and LLM-based).
We assume fixed-duration tasks that finish successfully.
We scale the number of goals and the number of robots for these tasks, and compute the percentage of idle time across combinations of task planners and allocators in
Table~\ref{tab:idle-times}.
The monolithic planner, which plans all goals jointly, leads to high idle time.
Introducing structure via a DAGs reduces idle time, with similar performance between both BigDAG and per-goal DAG. 
However, using the MILP-based allocator is generally leads to less idle time, with the least idle time achieved via the per-goal DAG planner.

We qualitatively found that Per-Goal DAG is effective when each goal is best completed in isolation, and when completing that goal requires most of the fleet to work on separate subtasks. 
This case leads to lower idle time, as each robot can independently work on separate subtasks with few dependencies across goals.
When the goals themselves happen to be more interconnected, it is more efficient to accomplish multiple subtasks at once. 
Thus, the Per-Goal DAG is not as efficient as Big-DAG when goals may share resources or dependencies. 
However, due to LLM context limits, Big-DAG may not perform well if there there are a large number of goals.
We find there are trade-offs to each task planning approach.

\subsection{Real Robot Experiments.}
\label{realdemo}
We also demonstrate real-world experiments with a Toyota HSR mobile manipulator and a LoCoBot navigation robot.
We focus on showcasing two demos, one where we show the planning capabilities of our system, and another where replanning is required to send new information to a different robot.

We design a simple world state with three locations: a kitchen, a hallway, and a living room, each defined by their \texttt{(x, y, yaw)} coordinates on each robot according to their 2D environment maps. 
In addition, we include simple information about the location of specific objects, the existence of other objects but not their locations, and the current location of the robot.

Additionally, we found that when using learned policies for manipulation, failures occurred regularly, causing frequent replanning. 
To focus our demos on the planning system, we teleoperated the manipulation phases of the subtasks to ensure more reliability.
As we design more robust manipulation policies, we also intend to more tightly integrate robust failure recovery and replanning depending on how the robot fails its task or if the robot discovers new information while completing a task. 

Our framework is capable of managing any kind of robotics stack on each robot as long as each robot can parse and execute any given language-defined subtask.
For each robot, we use code generation similar to ProgPrompt~\cite{singh2023progprompt} to parse a given subtask and generate code for solving a task.
For the navigation robot, we define functions for navigation, object detection via YOLOv8~\cite{yolov8}, and search. 
For the HSR mobile manipulator, we add additional manipulation tool calls for pick and place.
In the generated code, the robots are also able to propogate information back to the central planner if they find new information or if a failure requires replanning.

\textbf{Multi-Robot Planning Demonstration.} 
Given the HSR mobile manipulator and Locobot navigation robots, we provide two goals to our planner: ``Bring the first cup to the hallway" and ``Bring the second cup to the living room."
In addition to the world state described in the previous section, we described that the robots are in unknown locations and that the cups are in the kitchen as shown in Figure~\ref{fig:fig3}.
The DAG generated for this plan can be found in Figure~\ref{fig:fig4}.
We find reasonable task splits with dependencies on how the HSR should pick up the cup and place the cup on the LoCoBot's basket. 
In Figure~\ref{fig:fig3}, we show that the task execution successfully accomplishes both goals of bringing the cups to their respective rooms. 

\textbf{Replanning Demonstration.} 
Our planning system also supports replanning in the case of failure of a robot to complete a task or when new information is discovered in the environment by a robot that necessitates a new plan. 
In this demonstration, we design a single goal: ``Find a cup somewhere and bring it to the hallway."
To ablate only the replanning component, we manually assign the HSR the tasks of ``Go to the kitchen" followed by ``Search for the cup in the kitchen,", while we assign the LoCoBot to similarly search the living room. 
In this demonstration, the LoCoBot finds the cup in the living room and signals to the central planner about the new information.
A new plan is then generated involving the HSR going to the living room, picking up the cup, navigating to the hallway, and placing the cup in the hallway.
We intentionally defined the initial task plan to ensure the LoCoBot found the cup first as it does not have manipulation capabilities, and would require RobotFleet to correctly replan with the HSR.

\section{Conclusion}
In this work, we have presented RobotFleet, an open-source framework for centralized multi-robot task planning that unifies LLM-based planning, mixed-integer optimization, and containerized execution into a modular autonomy stack. Through a combination of structured DAG representations and flexible allocation strategies, RobotFleet enables heterogeneous robot fleets to efficiently decompose high-level goals into interdependent subtasks and distribute them across robots. We have also validated our system on real robotic platforms showcasing dynamic replanning and coordinated execution between a Toyota HSR manipulator and a LoCoBot navigator. By providing clear abstractions for planning, scheduling, and execution, RobotFleet lowers the barrier for researchers and practitioners to build scalable, open-source multi-robot applications.


\textbf{Limitations.} 
Currently, in the idle time experiments, RobotFleet assumes fixed task durations, which is typically unrealistic. 
We also assumed that every task was successful. The replanning module still requires significant improvements. We also teleoperated the manipulation tasks, though we plan to incorporate and use actual language-guided manipulation policies in the future.
Additionally, we currently only support specifying capabilities with natural language in YAML files and using those descriptions with LLMs and string parsing in other modules. In our demo robots' YAMLs, our capabilities were very broad in language terms, and in scenarios where capabilities are much more physically constrained and sensitive to real-time environment changes, simple language descriptions of capabilities may not be enough. More work will be required on robust capability and constraint representations.

Finally, our real robot experiments focus on two robots and predefined tasks; scaling to larger, more dynamic deployments will require further optimization of communication protocols and task planning.



\section*{Acknowledgments}


\bibliographystyle{plainnat}
\bibliography{references}

\end{document}